\def\year{2019}\relax
\documentclass[letterpaper]{article}
\usepackage{aaai19}
\usepackage{times}
\usepackage{helvet}
\usepackage{courier}
\usepackage{url}
\usepackage{graphicx}
\usepackage{diagbox}
\usepackage{multirow}
\usepackage{romannum}
\usepackage{amsmath}

\title{Who Blames Whom in a Crisis? Detecting Blame Ties from News Articles Using Neural Networks}
\author{Shuailong Liang, Olivia Nicol, Yue Zhang\\
		Singapore University of Technology and Design \\
        8 Somapah Road, Singapore 487372\\
		shuailong\_liang@mymail.sutd.edu.sg,
		\{olivia\_nicol, yue\_zhang\}@sutd.edu.sg
}

\begin{document}

\maketitle

\begin{abstract}
Blame games tend to follow major disruptions, be they financial crises, natural disasters or terrorist attacks. To study how the blame game evolves and shapes the dominant crisis narratives is of great significance, as sense-making processes can affect regulatory outcomes, social hierarchies, and cultural norms. However, it takes tremendous time and efforts for social scientists to manually examine each relevant news article and extract the blame ties (A blames B). In this study, we define a new task, Blame Tie Extraction, and construct a new dataset related to the United States financial crisis (2007-2010) from {\it The New York Times}, {\it The Wall Street Journal} and {\it USA Today}. We build a Bi-directional Long Short-Term Memory (BiLSTM) network for contexts where the entities appear in and it learns to automatically extract such blame ties at the document level. Leveraging the large unsupervised model such as GloVe and ELMo, our best model achieves an F1 score of 70\% on the test set for blame tie extraction, making it a useful tool for social scientists to extract blame ties more efficiently.
\end{abstract}

\section{Introduction}

Blame is an issue that has been receiving increasing attention in the social sciences in recent years ~\cite{alicke2000culpable,hobolt2014blaming,hood2010blame,shaver2012attribution}. In particular, more attention has been placed on blame dynamics following major disruptions, such as natural disasters ~\cite{malhotra2008attributing}, financial crises~\cite{nicolno,tourish2012metaphors}, and terrorist attacks~\cite{olmeda2008reversal}. Studying blame is of great significance as sense-making processes inform what and who a society values, and ultimately shape lawmaking. For instance, the intense blame targeting Wall Street during the financial crisis (2007-2010) helped lawmakers pass the July 2011 Dodd-Frank Wall Street Reform and Consumer Protection Act.

\begin{figure}[t!]
\centering
\includegraphics[width=\columnwidth]{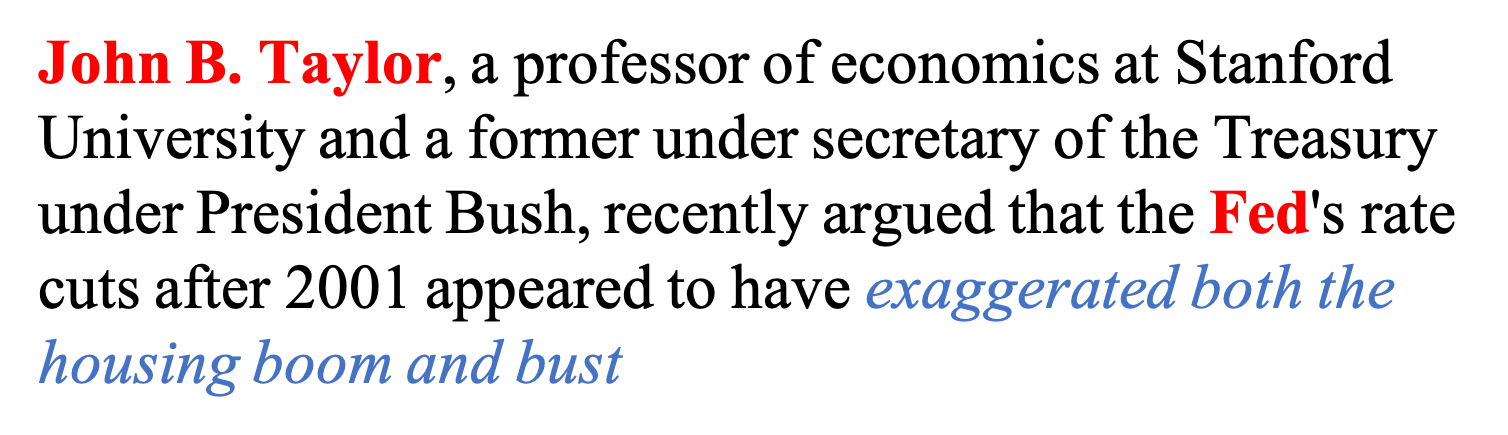}
\caption{An example sentence from our dataset containing a blame tie. The red/{\bf bold} words are entities involved in a blame tie, and the blue/{\it italic} words are supporting evidence that the blame tie exists.}
\label{figure:introdemo}
\end{figure}

Although the problem is important, it takes tremendous time and efforts for social scientists to manually examine each relevant news article and extract the blame ties (A blames B). In Figure~\ref{figure:introdemo}, for example, the tuple (John B. Taylor, Fed) is extracted as a blame tie. Recently, deep neural networks have proved very powerful at solving many social science problems~\cite{li-hovy:2014:EMNLP20141,rule2015lexical,bail2016combining}. Based on the dataset annotated from several new media excerpts on blame, we investigate automatic ways to extract blame ties from new articles.

There are three main challenges. First, some patterns only have blame meanings in specific contexts. For instance, {\it ``It was lenders that made the lenient loans, it was home buyers who sought out easy mortgages, and it was Wall Street underwriters that turned them into securities.''} ({\it The Wall Street Journal}, Aug 2007), only with the background of the financial crisis can we identify that the blamed targets are {\it lenders}, {\it home buyers} and {\it Wall Street}. Second, there are many ways to attribute blame and the structure of the sentences can be quite complex. Third, it is common for journalists to use metaphors and ironies to designate actors.

\begin{table*}[t!]
\centering
\small
\begin{tabular}{ | p{.5\columnwidth} | p{.5\columnwidth} | p{.25\columnwidth} | }
\hline
\multicolumn{3}{|l|}{\bf Articles } \\
\hline
\multicolumn{3}{ | p{.95\textwidth} | }{
...
Ordinarily, {\bf [Americans]} welcome lower interest rates. But many feel differently this time. Some think the economy is fine and inflation is the main danger. But a moral element is also at work: Many think a rate cut would reward foolish speculation and {\bf [Wall Street]} greed at the expense of the thrifty (1).

``The {\bf [Federal Reserve]} needs to stand its ground and not bail out hedge funds -- they should have known better to begin with!'' {\bf [Suzanne Mitchell]}, an administrative assistant at a Houston real-estate company, says in an email (2). In an interview, she adds: ``I'm very sorry that {\bf [people]} took out \$450,000 mortgages with no money down ... {\bf [people]} ought to be responsible for the loans they take out.'' (3)
\ldots

But Mr. {\bf [Brason]} contrasts that with the far greater reliance on borrowed money that is typical nowadays. Some ``of us ... weren't buying up five or 10 properties without any money down,'' he says. ``{\bf [People]} took the risks and should pay the price. (4) A lot of others at the higher end of the food chain, the {\bf [investment bankers]} and {\bf [hedge-fund managers]} were making oodles of fee-income money and frankly, there's a lot of public opinion that it was excessive. (5) ''
\ldots
} \\
\hline
\hline
{\bf Blame Source} & {\bf Blame Target} & {\bf Causality Link}\\
\hline
Americans($e_1$) & Wall Street($e_2$) & (1) \\
Suzanne Mitchell($e_3$) & Federal Reserve($e_4$) & (2) \\
Suzanne Mitchell($e_3$) & people($e_5$) & (3) \\
Brason($e_6$) & people($e_5$) & (4) \\
Brason($e_6$) & investment bankers ($e_7$) & (5) \\
Brason($e_6$) & hedge-fund managers ($e_8$) & (5) \\
\hline
\end{tabular}
\caption{An article titled {\it Rate Cut Has Foes on Main Street} ({\it The Wall Street Journal}, September 2007). Top: paragraphs of the article containing several blame patterns. The entities are in the brackets. Bottom: blame ties extracted from the article.}
\label{table:samples}
\end{table*}

\begin{table}[t!]
\centering
\small
\begin{tabular}{| c | c c c c c c c c | }
 \hline
 \diagbox{{\bf source}}{{\bf target}} & $e_1$ &  $e_2$  & $e_3$ & $e_4$ & $e_5$ & $e_6$ & $e_7$ & $e_8$ \\
 \hline
 $e_1$ & - & 1 & 0 & 0 & 0 & 0 & 0 & 0 \\
 $e_2$ & 0 & - & 0 & 0 & 0 & 0 & 0 & 0 \\
 $e_3$ & 0 & 0 & - & 1 & 1 & 0 & 0 & 0 \\
 $e_4$ & 0 & 0 & 0 & - & 0 & 0 & 0 & 0 \\
 $e_5$ & 0 & 0 & 0 & 0 & - & 0 & 0 & 0 \\
 $e_6$ & 0 & 0 & 0 & 0 & 1 & - & 1 & 1 \\
 $e_7$ & 0 & 0 & 0 & 0 & 0 & 0 & 0 & 0 \\
 $e_8$ & 0 & 0 & 0 & 0 & 0 & 0 & 0 & 0 \\
 \hline
\end{tabular}
\caption{Matrix representation of the blame ties in Table~\ref{table:samples}.}
\label{table:samplemat}
\end{table}

We design several neural models to address the problem. First, we leverage a neural network to learn the prior information about entities for blame tie extraction. In particular, a neural network is used to learn dense vector representations of entities so that similar entities can be visualized close to each other in the embedding space, and the likeliness of one entity to blame another can be inferred automatically without further knowledge. Second, we build a BiLSTM neural network to represent contexts, which can be used to predict blame ties between entities mentioned in the news articles using linguistic clues. Finally, a model that integrates entity knowledge and linguistic knowledge is constructed by integrating the two respective networks.

We conduct a case study on blame games for the U.S. financial crisis (2007-2010), the most important event since the Great Depression, which led to at least \$6 trillion in losses~\cite{luttrell2013assessing}. Results show that our model can effectively learn both entity knowledge and linguistic clues for blame ties. For example, it can successfully extraction entity relation with regard to blame ties from the crisis, such as the fact that both Wall Street and McCain tend to blame the same targets (Obama and Bernanke). In addition, the model can generalize to new cases of extracting blame patterns automatically. Our implementation and trained models are released at https://github.com/Shuailong/BlamePipeline.

\section{Related Work}
\label{task}

NLP has become increasingly popular in the social science area. \citeauthor{o2010tweets} [\citeyear{o2010tweets}] aligned sentiment measured from Twitter to public opinion measured from polls, and found the two correlate well. \citeauthor{bamman2015open} [\citeyear{bamman2015open}] tried to use text data to estimate the political ideologies of individuals. \citeauthor{mohammad-EtAl:2016:SemEval} [\citeyear{mohammad-EtAl:2016:SemEval}] create the SemEval 2016 Task 6 called Stance Detection Task, which detects the Twitter user's stance towards a target of interest. \citeauthor{preoctiuc2017beyond} [\citeyear{preoctiuc2017beyond}] also predicted the political ideologies of Twitter users, in a more fine-grained form. Social scientists used NLP along with network analysis etc. to analyze social media texts~\cite{rule2015lexical} and the State of Union addresses in United States~\cite{bail2016combining}.

The Blame Tie Extraction task can be regarded as a special case of relation extraction~\cite{miwa2016end}. Relation extraction solves the task of classifying a pair of entities into one of several pre-defined categories, such as Cause-Effect and Component-Whole~\cite{hendrickx2009semeval}, while the Blame Tie Extraction task requires extracting all the blame ties among the entities of interest in an article. Our work differs from existing work on relation extraction in two main aspects. First, our work is at the document level and the data is sparse, while most existing work on relation extraction focuses on the sentence level~\cite{nguyen2015relation}. Second, our work explicitly uses entity prior information on blame patterns, which does not make sense in general domain relation extraction. Most existing work mixes entity and content information in modeling relations.

In blame game research, social scientists care more about a few key players instead of all the entities, and most entities in the passage are irrelevant for studying the blame game~\cite{nicolno}. Therefore, in this paper, we assume that the entities of interest are already given, and we only need to extract blame ties from these entities.

To the best of our knowledge, our work is the first to study the Blame Tie Extraction task using NLP techniques.

\section{Dataset}
\label{dataset}

We manually create a dataset on the U.S. financial crisis. The dataset is drawn from three newspapers in the U.S., including {\it The New York Times}, {\it The Wall Street Journal} and {\it USA Today}, chosen for three main reasons. First, they are the most widely circulated newspapers in the United States. Second, they cover the social spectrum from elite to mass. Third, they also cover the political spectrum: from the quite liberal {\it New York Times}, to the centrist {\it USA Today}, to the conservative {\it Wall Street Journal}~\cite{gentzkow2010drives,groseclose2005measure}. The time period studied here spans from August 2007, after the first warning signs for the crisis appeared, to June 2010, right before the signature of the largest set of financial regulations since the Great Depression.

We use a set of keywords to filter the articles gathered from Factiva\footnote{https://www.dowjones.com/products/factiva/} and LexisNexis\footnote{https://www.lexisnexis.com}, getting articles containing blame patterns for the crisis. The keywords we use are blame-related ({\it attack}, {\it accuse}, {\it misconduct}, \ldots) or event-related ({\it financial crisis}, {\it global recession}, {\it housing bubble}, \ldots)\footnote{The full list of keywords are released along with the code.}. There are in total 70 blame-related keywords and 13 event-related keywords (stem form). The two classes of keywords are combined together to filter the articles.

Blame incidences are manually coded for each article. The identification of a blame pattern requires the presence of 1) a blame source, 2) a blame target, and 3) a causality link. An example of a sentence containing a blame instance would be as follows: ``{\it Sen. Richard Shelby, R-Ala. [Blame Source]\ldots said the FED [Blame Target] `kept interest rates too low for too long, encouraging a housing bubble and excessive risk taking'[causality link]}.'' ({\it USA Today}, December 2009). The blame source and target form a blame tie. Table~\ref{table:samples} gives an example article and its annotations and Table~\ref{table:dataset} shows the statistics of the dataset with the number of blame ties.

To ensure the reliability of the dataset, we ask two more annotators to annotate a subset of the dataset. Specifically, we sample 100 articles, including 13 articles from {\it USA Today}, 43 articles from {\it The New York Times}, and 44 articles from {\it The Wall Street Journal}, in proportion to the respective number of articles of the three newspapers in the whole dataset. Then we run evaluations using the two annotators' results against the gold data. The average of the F1 score of the two annotators is 94.425\%, and the Fleiss's kappa is 0.8744, which illustrates the strong inter-annotator agreement of the dataset.

\begin{table}[t]
\centering
\begin{tabular}{l c c c}
 \hline
 & {\bf USA} &  {\bf NYT}  & {\bf WSJ} \\
 \hline\hline
 days & 310 & 736 & 648 \\
 articles & 132 & 429 & 438 \\
 blame ties & 353 & 787 & 754 \\
 \hline
\end{tabular}
\caption{Dataset size for the three newspapers. USA: {\it USA Today}. NYT: {\it The New York Times}. WSJ: {\it The Wall Street Journal}.}
\label{table:dataset}
\end{table}

In the training process, the annotated blame ties serve as positive samples. The negative samples are generated by removing the positive entity pairs from all possible permutations of the entities of interest. The sample statistics about the whole dataset are shown in Table~\ref{table:samplestats}. Theoretically, the number of negative samples increases quadratically with the number of entities in the article. In our dataset, the average number of entities we consider per article is 3. Therefore the dataset is rather balanced.

\section{Task}

We formulate the Blame Tie Extraction task as follows. Given a news article $d$ and a set of entities $e$, we have $ |e| \cdot (|e| - 1)$ possible directed links among them. We assign label $1$ to a pair ($s$, $t$) when entity $s$ blames entity $t$ based on article $d$, otherwise we assign label $0$ to the pair. We can use a matrix for a more intuitive illustration. For the example in Table~\ref{table:samples}, the matrix constructed is shown in Table~\ref{table:samplemat}.

For a given entity pair $(s, t)$ with label $l$, we would like to maximize the likelihood $$L = P(l | s, t, d), l \in \{0, 1\}$$

In order to predict whether a blame tie exists between two entities based on the article, we have two sources of information to utilize. One is the entities themselves. For instance, we know that Democrats tend to blame Republicans so as to weaken their political opponents, and tend to blame Wall Street so as to gain popular support to impose a stringent set of financial regulations. The other one is the contexts in which the entities are mentioned. We rely on linguistic patterns of sentences to extract the blame ties. For instance, in this sentence ``{\it Who is to blame? Hedge funds, for one, he says.}'' ({\it The Wall Street Journal}, Sept 2007), the linguistic structure indicates that the entity appearing after the question is the blame target entity, and the narrator is the blame source entity.

\begin{table}[t]
\centering
\begin{tabular}{l c}
 \hline
 number of articles & 998 \\
 number of samples & 8562 \\
 number of entities/article & 2.97 \\
 average neg/pos ratio per article & 2.19 \\
 total neg/pos ratio & 3.61 \\
 \hline
\end{tabular}
\caption{Sample statistics.}
\label{table:samplestats}
\end{table}

\section{Models}

\begin{figure*}[t!]
  \centering
  \includegraphics[width=\textwidth]{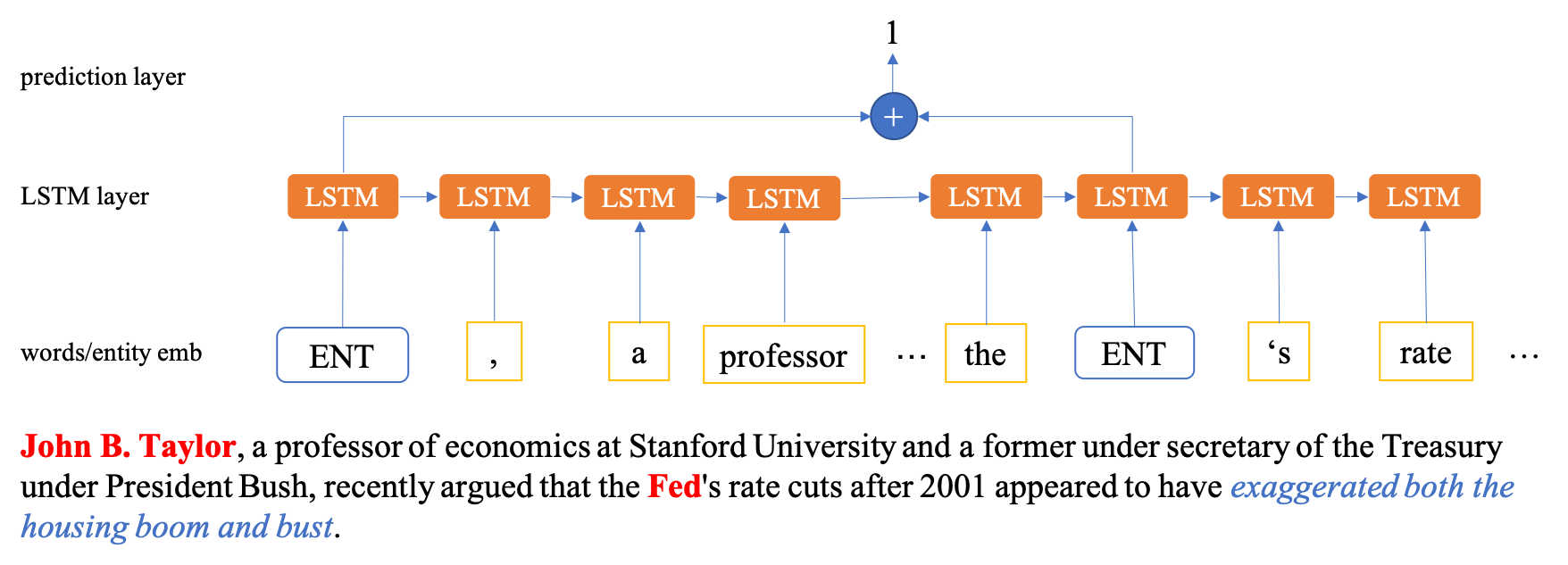}
  \caption{Context Model. LSTM is used to encode the sentences, and the hidden vectors at the positions of the entity are pooled together into a single vector to represent the context of the entity. The source entity context vector and the target entity context vector are concatenated together to be sent to the prediction layer. The prediction layer predicts 1 if the source to target blame tie exists otherwise 0.}
  \label{figure:contextmodel}
\end{figure*}

We first introduce a simple rule-based model. Next, we introduce three neural models. The Entity Prior Model which directly extracts prior information about entities (i.e.\ which entities tend to blame which entities). The Entity Prior Model can learn entity information from its blame patterns such as political standing, and it can generalize to new events with the same entities. Then, we mask out the entity mentions from the text, relying purely on the contexts surrounding the mentions of the entities. The Context Model can generalize to new entities and across historical periods. Finally, a model combining the two is built to investigate the interactions.

\subsection{Rule-based Model}
As a baseline model, we use a simple rule to decide if a blame tie exists between two entities: if the minimal sentence distance between the two entities are less than or equal to $d$, AND a blame related keyword appears in any of the sentences mentioning either of the two entities, we determine that a blame tie exists between the two entities. According to the distribution of the minimal distance between two entities in a blame tie in our dataset, over 90\% of the blame entity pairs has a minimal sentence distance less than or equal to 3. Therefore, we set $d=3$.

To determine the direction (A blames B or B blames A), we define the {\bf aggressiveness} of the entity, which is the percentage of the entity being the blame source among all the blame ties related to this entity in the training data. The entity with higher aggressiveness is the blame source entity, and the other one is the blame target. When there is a tie, we use random guess. When the entity is unknown, we use 0.5 as the aggressiveness score.

\subsection{Entity Prior Model}
\label{entitymodel}
We use the fully connected feedforward neural network (FCN) to collect entity prior knowledge, namely who is likely to blame whom without additional information. To elaborate, we represent entities by their embeddings, concatenate the embeddings of the blame source and target entity, and then stack a fully connected layer to learn the interactions between the entities. The FCN outputs the probability of the blame tie between the source entity and the target entity: $$f_{score} = ([\mathbf{E}^e(e_s); \mathbf{E}^e(e_t)]) \cdot W^e + b^e $$

$e_s$ and $e_t$ represent the source entity and target entity index, respectively, $\mathbf{E}^e$ is the embedding matrix for entities, `;' is the concatenation operator, and $W^e$ and $b^e$ are parameters. Specifically, $\mathbf{E}^e \in \mathbf{R}^{n\times m}$, $n=707$ is the number of entities in training set, $m$ is the entity embedding dimension, which is tuned as hyperparameters. $W^e \in \mathbf{R}^{2m\times 2}$ and $b^e \in \mathbf{R}^{2} $. $\mathbf{E}^e$ is initialized with standard normal distribution. At test time, we use the $\langle$UNK\_ENT$\rangle$ to indicate unknown entities from the training set.

Like word embedding, we hope to learn meaningful representations of the entities, i.e., the entities sharing similarity blame behavior patterns will stay close in the embedding space.

\subsection{Context Model}
\label{contextmodel}

The Entity Prior Model tells how likely a particular entity is to blame or be blamed without further information. In contrast, the Context Model relies only on linguistic clues from news articles, finding blame patterns explicitly or implicitly mentioned. The Context Model is thus useful across different political settings where the entities are different.

To model context information about an entity, we first locate the positions of all occurrences of the entity in the article. A position is represented by a tuple $(i, j)$, where $i$ and $j$ denotes sentence number and word number, and the positions of blame source and blame target are denoted as $pos^s$ and $pos^t$, respectively. Second, we replace each entity mention by a special $\langle$ENT$\rangle$ token, so that we do not have any information regarding the entity itself. Third, we run a bidirectional LSTM on sentences containing the entity and use the LSTM output to represent the context of each word:

$$h^i_j=\mathrm{LSTM}(\mathbf{E}^w(w^i_j))$$

$w^i_j$ is the $j$-th word of the $i$-th sentence, $\mathbf{E}^w$ is the embedding matrix for words, and $h^i_j$ is the concatenation of the LSTM outputs of the last layers from both directions. If $(i, j) \in pos^s$ or $(i, j) \in pos^t$, $h^i_j$ will be used to represent the context of entity $s$ or $t$.

Since an entity may appear at multiple positions in one article, we may have multiple representations of the entity context. Pooling is used to reduce the embeddings into one single vector. The pooling result of the contexts representations of source and target entity are denoted as $V_s$ and $V_t$, respectively: $$V_e = \underset{(i, j) \in pos^e} {\mathrm{pool}} (h^i_j), e \in \{s, t\}$$where $\mathrm{pool}$ denotes the pooling function, for which we try random selection, mean, max and attention. For attention pooling, we use a two-layer feedforward neural network to score each vector representations, using softmax function to normalize them as weights, and use the weighted sum of the representations as the entity representation. Once we obtain representations of both the source and target entity, we concatenate them and then use a fully connected layer to learn a score of this entity pair: $$f_{score} = ([V_s; V_t]) \cdot W^v + b^v $$ $W^v \in \mathbf{R}^{2H \times 2}$ and $b^v \in \mathbf{R}^{2}$ are both parameters. $H$ is the dimension of BiLSTM outputs. The model architecture is depicted in Figure~\ref{figure:contextmodel}.

\subsection{Combined Model}

To incorporate the prior information about entities, we concatenate two additional vectors on the basis of the Context Model, which are the embeddings of the blame source entity $\mathbf{E}^e(s)$ and blame target entity $\mathbf{E}^e(t)$ from section \ref{entitymodel}: $$f_{score} = ([\mathbf{E}^e(e_s); V_s; \mathbf{E}^e(e_t); V_t]) \cdot W^c + b^c $$ $W^c \in \mathbf{R}^{2H + 2m}$ and $b^c \in \mathbf{R}^{2}$ are both parameters. We expect the model to simultaneously learn how to extract blame ties from context, and also learn the representations of the entities themselves as a byproduct.

\section{Experiments}
\label{experiment}

We conduct experiments on our dataset using the Entity Prior Model, the Context Model, and the Combined Model and compare the performance among different models. To find the best model architecture, we also conduct development experiments to investigate the effects of different pooling functions and pretrained word embeddings.

\subsection{Experimental Settings}

\begin{table}[t!]
\centering
\begin{tabular}{l c c c}
 \hline
 &{\bf min} & {\bf avg}  &{\bf max} \\
 \hline\hline
sentences per doc & 4 & 45 & 384 \\
words per sentence & 1 & 26 & 159 \\
words per doc & 133 & 1,209 & 9,064 \\
 \hline
\end{tabular}
\caption{Sentence and words statistics.}
\label{table:lengthstatistic}
\end{table}

Stanford CoreNLP ~\cite{manning-EtAl:2014:P14-5} is used to tokenize articles into sentences and words. The sentence length statistics of the dataset articles are shown in Table~\ref{table:lengthstatistic}. There is no further preprocessing except that words are converted to lower case. The dataset is split into train, dev, and test set at the document level by the ratio of 8:1:1. We use F1 on the positive class to measure the performance of the model.

To investigate whether the learned entities representations are helpful in blame tie extraction, we conduct another round of evaluations on the known entities of dev and test set as shown in Table~\ref{table:finalresult}. KNOWN denotes entity pairs both of which appear in the training data, while ALL denotes all entity pairs. By comparing the model performance on KNOWN with that of ALL, we can evaluate the usefulness of entities representations. Conversely, we can also evaluate the robustness of the model against unknown entities.

Leveraging large unsupervised data has proved to be helpful for many NLP tasks, especially for tasks with small datasets. Word2Vec~\cite{DBLP:journals/corr/abs-1301-3781} or GloVe~\cite{pennington2014glove} can be used to pretrain word embeddings on larger external dataset. ELMo~\cite{Peters:2018} word vectors are internal states of a deep bi-directional language models, and can effectively capture the syntax and semantics of words. We conduct experiments on GloVe and ELMo, and investigate the effects of these pretrained word embeddings.

Our models are implemented in PyTorch\footnote{http://pytorch.org}. During training, Adam~\cite{kingma2014adam} is used as the optimizer, and the default learning rate is adopted. We use the minibatch size of 50 for all three models. Dropout~\cite{hinton2012improving} of $0.5$ are used to prevent overfitting. Dropout is applied to word embeddings and RNN outputs. We set gradient clipping to 3 to stabilize the training process. The maximum number of epochs is 30 and we use early stopping techniques with a patience of 10 epochs. For the model hyperparameters, word embedding size is $100$, LSTM hidden size is $100$ for each direction, and entity embedding size is $50$.

\subsection{Development Experiments}

Before turning to the BiLSTM model, we use the small window before and after the entity to extract the context information. Formally, we assume that one entity appears only once in a sentence. Given an entity $e$ appearing at sentence $s$, if the index of $e$ is $i$, and window size is $w$, we use concatenation of the embeddings of words $s_{i-w}\ldots s_{i-1}$ and $s_{i+1}\ldots s_{i+w}$ as the context of $e$. We call this model Bag of Embeddings (BoE). We try the window sizes of 3 and 6 and report the higher dev result. If an entity appears multiple times in an article, we randomly select one representation (BoERand).

As stated in the Context Model section, we compare different pooling methods used to aggregate multiple contexts representations, and bi-directional LSTM with a single forward LSTM. The results of ALL dataset are shown in Table~\ref{table:contextresult}. BiLSTM works better than LSTM for every pooling method since we can take advantage of information from both directions of the sentence. Random pooling has the worst result, mainly because a lot of important information is lost. Max pooling has the highest score on most cases, therefore we use BiLSTM + max pooling in the following experiments.

In comparison to BoE models, LSTM models are worse. The reason may be that BoE models consider contexts from both sides. BiLSTM models perform better than BoE models since the former can model more context and preserve word order information.

\begin{table}[t]
\centering
\begin{tabular}{l c | l c}
\hline
{\bf Model} & {\bf dev F1 } & {\bf Model} & {\bf dev F1} \\
\hline\hline
BoERand & 54.60 && \\
\hline
LSTMRand & 50.51 & BiLSTMRand & 56.43 \\
LSTMMean & 50.15 & BiLSTMMean & 61.95 \\
LSTMMax & 51.49 & BiLSTMMax  & 62.26 \\
LSTMAttn & 50.17 & BiLSTMAttn & 61.92 \\
\hline
\end{tabular}
\caption{Experiment results of Context Model using different pooling functions.}
\label{table:contextresult}
\end{table}

\begin{table}[t]
\centering
\begin{tabular}{l c c}
\hline
{\bf Model}  & {\bf dev F1 } & {\bf test F1 } \\
\hline\hline
random  &  62.26  & 56.11 \\
GloVe fixed  & 63.09  & 62.10 \\
GloVe tuned  & 61.37  & 57.75  \\
ELMo  & {\bf 73.16} & {\bf 66.35} \\
\hline
\end{tabular}
\caption{Experiment results of Context Model using different pretrained word vectors.}
\label{table:pretrain}
\end{table}

To investigate the effect of pretrained word embeddings, we initialize the word embedding parameters with random initialization, GloVe pretrained word embeddings, and ELMo models. Since the official release of pretrained ELMo model has the output dimension of 1024, we do a linear transformation to reduce the output dimension to 100. The transformation matrix is part of the model parameters and are tuned during training. For GloVe, we use fixed and tuned versions of embeddings. For ELMo model, since it slows down the training significantly, we do not tune the model parameters. The results using BiLSTM+Max pooling are shown in Table \ref{table:pretrain}. Fixed GloVe vectors improve the F1 score on dev and test set by 0.83\% and 4.99\%, respectively. The tuned version of GloVe does not improve that much, due to the fact that the dataset is small and too many parameters will cause overfitting. The pretrained ELMo model improves the F1 on the dev and test set by 10.90\% and 10.24\%, respectively, compared with random initialization, proving the powerfulness of ELMo model.

\subsection{Results}

Table~\ref{table:finalresult} details the final result of the Entity Model, Context Model, and Combined Model. For comparison, we also include a baseline of random guessing and rule-based model. For the Context Model, we use the BiLSTM+Max pooling and use ELMo model to obtain words representations.

\begin{table}[t]
\centering
\begin{tabular}{l | c c | c c}
\hline
\multirow{2}{*}{\bf Model} & \multicolumn{2}{c|}{\bf KNOWN}  & \multicolumn{2}{c}{\bf ALL} \\
 & {\bf dev F1 } & {\bf test F1} & {\bf dev F1 } & {\bf test F1 } \\
\hline\hline
random guess & 38.81 & 38.04 & 37.39 & 32.96 \\
\hline
rule-based & 69.14  & 58.97 & 70.45 & 61.54 \\
\hline\hline
entity & 73.97 & {\bf 70.97} & 61.07 & 60.06 \\
\hline
context & 74.29  & 63.11 & 73.16 & 66.35 \\
\hline
combined & {\bf 81.75}  & 68.67 & {\bf 76.13}  & {\bf 69.92} \\
\hline
\end{tabular}
\caption{Experiment results of baseline models and three proprosed models on KNOWN data and ALL data.}
\label{table:finalresult}
\end{table}

Our rule-based model outperforms random guess at a large margin. Surprisingly, it achieves a higher result than the Entity Model on ALL data. The reason is that the Entity Model cannot generalize to new entities, therefore performs badly on ALL data with many unknown entities. Compared with the Entity Model on the KNOWN data, Entity Model performs better.

For the \textbf{Entity Model}, the performance on KNOWN entities is better than that on ALL entities. This result illustrates that the model learns prior information about the entities in the train set. From a visualization of entity embeddings using tSNE~\cite{maaten2008visualizing}, we find that entities with similar political backgrounds tend to be close to each other in the embedding vector space. For example, Wall Street and McCain are close to each other, which is intuitive since they were both blamed by Obama, and they both blamed Obama and Bernanke, according to the training data. In addition, Obama, Congress, and Bernanke also form a cluster, since they were all blamed by McCain and blamed the Fed.

For the \textbf{Context Model}, we can see that it can effectively extract blame ties from news articles without prior entity knowledge. The F1 on the KNOWN test set is 63.11\%; while on ALL test set, the F1 does not decrease as in the Entity Prior Model, it even increases to 66.35\%. Unlike knowledge about entities, linguistic knowledge generalizes robustly to unseen test data, where most entities do not exist in the training data. This implies that our model can be used to extract blame ties in other occasions where the political settings are highly different. For example, when the President of the United States changes, we can still use our model to predict how the new President will play the blame game.

For the KNOWN entities, the \textbf{Combined Model} does \textit{not} perform better than Entity Model. This shows that entities information alone may be useful in extracting blame ties than using contexts. However, such information could not be used when the entities are new, for instance, our ALL data. On the ALL data, the Combined Model achieves the best result on dev and test set, showing that the model can integrate context information as well as the entity prior information to make better predictions.

Therefore, the Context Model is the most robust one, applicable to extract blame tie among new entities. Entity prior information is helpful. If available, it can be used to boost the model performance.

\begin{table}[t!]
\centering
\begin{tabular}{| l | c | c | }
\hline
& {\bf CTX Wrong} & {\bf CTX Correct}\\
\hline
{\bf CMB Wrong} & 0/0 & 31/19 (\Romannum{2}) \\
\hline
{\bf CMB Correct} & 98/127 (\Romannum{1}) & 413/406 \\
\hline
\end{tabular}
\caption{Model comparison between Context Model (CTX) and Combined Model (CMB) on dev/test sets.}
\label{table:comparison}
\end{table}

\begin{table*}[t]
\centering
\small
\begin{tabular}{ | p{.95\textwidth} | }
\hline
{\bf Articles} \\
\hline
{\bf Trump} Accuses {\bf Russia} of Helping {\bf North Korea} Evade Sanctions (1)

President {\bf Donald Trump} accused {\bf Russia} in unusually harsh terms of helping {\bf North Korea} evade United Nations sanctions intended to press the country to give up its nuclear and ballistic missile programs. (2)

`` {\bf Russia} is not helping us at all with {\bf North Korea},'' {\bf Trump} said in an interview with Reuters on Wednesday (3). ``What China is helping us with, {\bf Russia} is denting. In other words, {\bf Russia} is making up for some of what China is doing.''

{\bf Trump} has leaned on China to curb its support for {\bf North Korean} leader Kim Jong Un's regime, and in exchange has so far laid off the punishing trade measures he promised against the U.S.'s largest creditor during his campaign.

North Korea's weapons programs are {\bf Trump}'s most urgent foreign crisis (4). He has vowed not to allow the country to develop a missile capable of carrying a nuclear warhead to the U.S. mainland, threatening war to prevent it if necessary. But Kim has plunged ahead, and his government made rapid advances with both its missile and nuclear technology after {\bf Trump} took office.

{\bf Trump}'s criticism of {\bf Russia} (5) is striking because members of Congress have said in the past that he was too reluctant to criticize {\bf Russia}'s foreign policy and too eager to establish good relations with President Vladimir Putin.
\\
\hline
\hline
{\bf Major Entities Mentioned}\\
\hline
Trump ($e_1$), Russia ($e_2$), North Korea ($e_3$)\\
\hline
\end{tabular}
\caption{An article from {\it Bloomberg} published on January 18, 2018. Top: paragraphs of the article containing blame patterns. The blame entities are in {\bf bold face}. Bottom: major entities appearing in the article.}
\label{table:news}
\end{table*}

\begin{table}[t]
\centering
\begin{tabular}{| c | c c c|}
 \hline
 \diagbox{{\bf source}}{{\bf target}} & $e_1$ &  $e_2$  & $e_3$\\
 \hline
 $e_1$ & - & 0.82 & 0.91\\
 $e_2$ & 0.02 & - & 0.40\\
 $e_3$ & 0.00 & 0.01 & -\\
 \hline
\end{tabular}
\caption{Prediction result of the article in Table~\ref{table:news} using the Context Model. The number represents the probability that the blame ties exists between the two corresponding entities.}
\label{table:newsresultcontext}
\end{table}

\section{Analysis}

To verify our hypothesis about why the Combined Model is better than the Context Model, we take several examples on which the Combined Model succeeds while the Context Model fails and vice versa. To evaluate the practicality and generalization of the model, we use our trained model to extract blame ties from several recent news articles.

\subsection{Case Study}

To analyze the fine-grained differences between behaviors of the Context Model and the Combined Model, we evaluate the two models on samples of the test data, and divide the samples into four classes, as shown in Table~\ref{table:comparison}.

Class \Romannum{1} samples are those on which the Context Model fails while the Combined Model works correctly. These samples usually involve entities that appear frequently in the train set. For instance, we want to figure out whether {\it Obama} blames {\it Republicans} based on the article titled {\it Obama Issues Sharp Call for Reforms on Wall Street} ({\it The New York Times}, April 2010). From sentences such as ``{\it The president and his allies have eagerly portrayed Republicans as handmaidens of Wall Street... Obama avoided incendiary language attacking Republicans...}'' we can see that the blame tie (Obama, republicans) holds. The Context Model predicts that the blame tie exists with a confidence score of 0.45, while the Combined Model gives a score of 0.97. The reason may be that the Combined Model learns prior information about Obama and Republicans: Obama blamed Republicans before in the training data, which makes sense since he is a Democrat.

Class \Romannum{2} samples are those on which the Combined Model fails, while the Context Model works correctly. In the {\it USA Today} article {\it Obama tells Wall Street to join in} (April 2010), the context is {\it ``...The president said the financial crisis, which has cost more than 8 million jobs so far, was `born of a failure of responsibility from Wall Street all the way to Washington.'...''}. The Context Model predicts that Washington does not blame Wall Street, at the confidence level of 0.62, which is true. However, the Combined Model mistakenly predicts a probability of 0.88 that the blame exists. This is because the Combined Model overly relies on entity information. Nevertheless, in Table~\ref{table:comparison} this class of samples is much fewer than class \Romannum{1} samples.

\subsection{Generalization to New Cases}

To validate the generalization of our model, we conduct further analysis on news articles beyond the time frame of the financial crisis. Since most entities in this new dataset do not appear in our financial crisis dataset, we use the context model for generalizability test. In particular, we manually annotate 13 recent articles containing 14 blame ties from Google News, mostly in January 2018, and use our pretrained Context Model to extract blame ties from the articles. The F1 on individual blame ties is 72.00\% on this new test data, and 8 out of 13 articles are labeled correctly. The result is consistent with the result of our financial crisis test set. This further demonstrates that the linguistic patterns for blame are generalizable to new scenarios. The articles on which the model fails mainly contain blame patterns that have not be seen in our financial crisis dataset.

Table~\ref{table:news} shows one of these new articles from Bloomberg Politics. In this news, United States Presidents accused Russia of helping North Korea evade United Nations sanctions. The blame tie is reflected in the sentences (1) (2) (3) and (5). From (4) we can infer that Trump has a negative attitude towards North Korea because of the nuclear weapons. The results of the Context Model are shown in Table~\ref{table:newsresultcontext}. The model successfully identifies the blame ties from Trump to Russia and Trump to North Korea, with a high confidence score, while the probabilities for the other entity pairs are low.

\section{Conclusion}

We investigated blame analysis, for which previous research looked at the evolution of blame frames, without systematically connecting the frames to the actors who produce them. Experiments show that neural networks are effective for identifying blame ties from news articles.  Our approach can enable researchers to quantify the importance of a frame and to understand how and why it became prevalent. It can also enable researchers to study actors' position alignments over time. To facilitate such research, we release our code and model for automatic blame tie extraction.

\section{Acknowledgments}
We thank Zhiyang Teng and Jie Yang for a lot of helpful discussions, Yan Zhang for helping with the inter-annotator agreement evaluation, Yu Yuan and Gokayaz Gulten and my other colleagues for helping with the proofreading. Yue Zhang is the corresponding author.

\bibliography{AAAI-LiangS.2972}
\bibliographystyle{aaai}

\end{document}